\pdfoutput=1

\documentclass[11pt]{article}

\usepackage[]{ACL2023}
\usepackage{multicol}

\usepackage{times}
\usepackage{latexsym}
\usepackage{hyperref}
\usepackage{epsfig}
\usepackage{amsmath}

\usepackage[T1]{fontenc}
\usepackage{tablefootnote}
\usepackage[utf8]{inputenc}

\usepackage{microtype}

\usepackage{soul}
\usepackage{inconsolata}

\title{Investigating Forgetting in Pre-Trained Representations Through Continual Learning}

\author{
    Yun Luo \textsuperscript{\rm1},
    Zhen Yang \textsuperscript{\rm2},
    Xuefeng Bai \textsuperscript{\rm1},
    Fandong Meng \textsuperscript{\rm2},
    Jie Zhou  \textsuperscript{\rm2},
    Yue Zhang \textsuperscript{\rm1,3}\footnote{Corresponding Author}
    \\
    \textsuperscript{1} School of Engineering, Westlake University, Hangzhou, 310024, P.R. China. \\
    \textsuperscript{2} Pattern Recognition Center, WeChat AI, Tencent Inc, Beijing, China.  \\
    \textsuperscript{3} Institute of Advanced Technology, Westlake Institute for Advanced Study, Hangzhou, 310024, P.R. China.  \\
    \texttt{\{luoyun, baixuefeng, zhangyue\}@westlake.edu.cn}\\
    \texttt{\{zieenyang, fandongmeng, withtomzhou\}@tentent.com}
}
    
\begin{document}
\maketitle
\begin{abstract}

Representation forgetting refers to the  drift of contextualized representations during continual training. Intuitively, the representation forgetting can influence the general knowledge stored in pre-trained language models (LMs), but the concrete effect is still unclear. In this paper, we study the effect of representation forgetting on the generality of pre-trained language models, i.e. the potential capability for tackling future downstream tasks. Specifically, we design three metrics, including overall generality destruction (GD), syntactic knowledge forgetting (SynF), and semantic knowledge forgetting (SemF), to measure the evolution of general knowledge in continual learning. With extensive experiments, we find that the generality is destructed  in various pre-trained LMs, and syntactic and semantic knowledge is forgotten through continual learning. Based on our experiments and analysis, we further get two insights into alleviating general knowledge forgetting: 1) training on general linguistic tasks at first can mitigate general knowledge forgetting; 2) the hybrid continual learning method can mitigate the generality destruction and maintain more general knowledge compared with those only considering rehearsal or regularization.

\end{abstract}

\section{Introduction}
Continual learning, which aims at training a model from a sequence of tasks, has attracted increasing attention in natural language processing (NLP) \cite{WUTONG}. It is a new learning paradigm  that aims to imitate the human capacity for continual learning and knowledge accumulation without forgetting previously learned knowledge, and also transfer the knowledge to new tasks to learn them better \cite{ke2022continual}. A typical application scenario for continual learning in NLP involves fine-tuning pre-trained language models (LMs) \cite{devlin2019bert,liu2019roberta,radford2019language} on various tasks, thus the rich information in the pre-trained representations can be shared and leveraged by different tasks.
However,  catastrophic forgetting (i.e., the phenomenon that the model's performance on preceding tasks severely decreases when the model is trained on a new task) becomes a salient challenge for making full use of the pre-trained representations.
Previous studies have shown that catastrophic forgetting is highly related to the forgetting of the contextualized representation (i.e. representation forgetting or representation drift) \cite{li2017learning,kirkpatrick2017overcoming,yu2020semantic,jie2022alleviating}.  


\begin{figure}
    \centering
    \includegraphics[width=0.9\hsize]{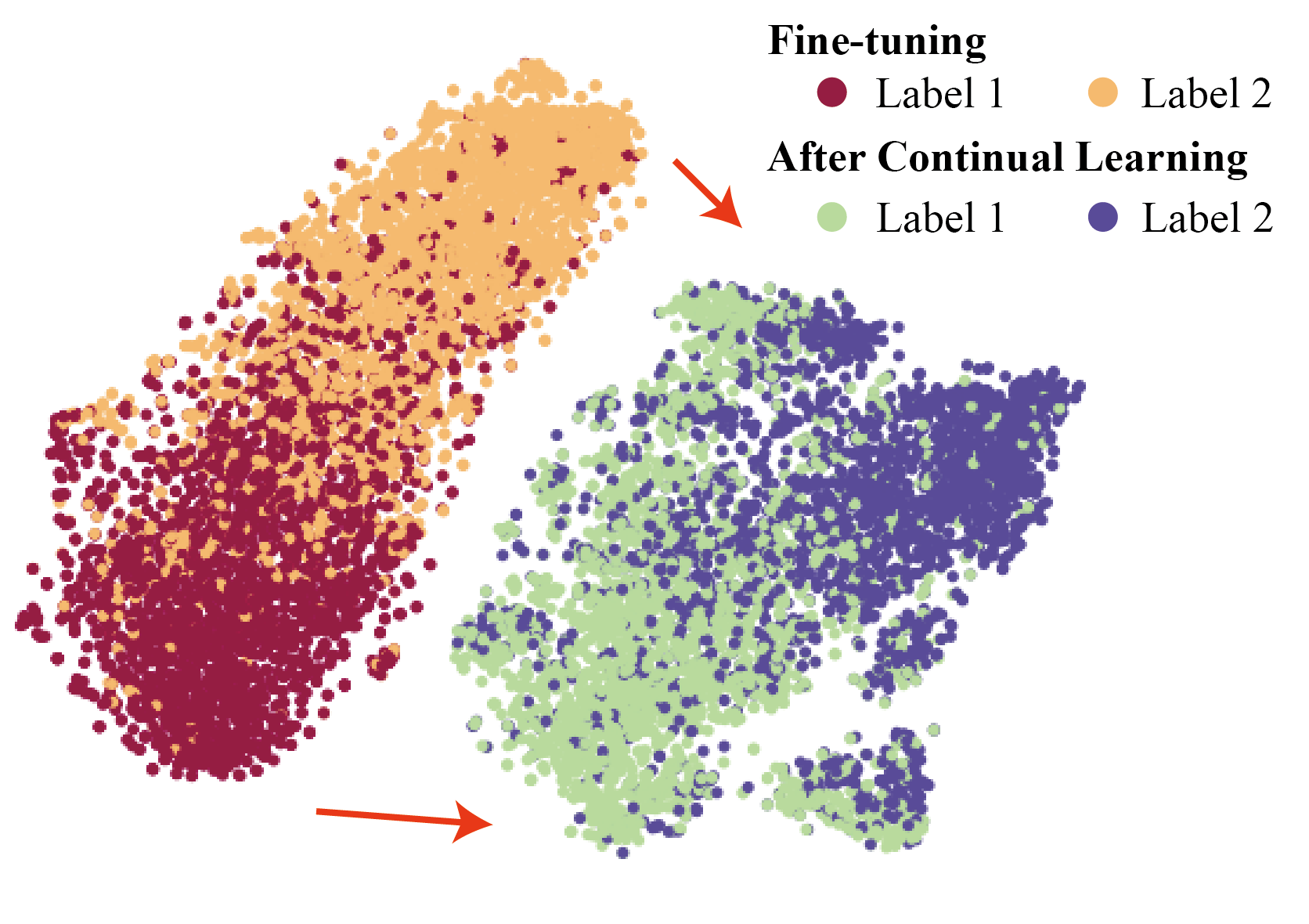}
    \caption{t-SNE visualization of the representations of natural language inference 
(NLI) data in our experiment. The representations by fine-tuning on NLI are shown in orange and red, and  they drift to the position of green and blue after continual learning. }
    \label{vis}
    \vspace{-5mm}
\end{figure}


Figure \ref{vis} presents an example of representation forgetting in the pre-trained LM. In Figure \ref{vis}, the contextualized representations for the natural language inference \cite{wang2018glue} task before and after the continual learning are illustrated. The BERT classification model is based on the continual learning strategy of experience replay \cite{de2019episodic} and the performance on the NLI task decreases from 83.57\% to 81.08\% after continual learning. Although performance is mainly maintained,  the representations obviously drift a significant margin in the feature space. When the model is continually trained on the task of sentiment analysis, its performance for sentiment classification is 60.56\%, 2.13\% lower than that solely trained on  the initial pre-trained BERT, which implies the effectiveness of pre-trained representations for downstream tasks has changed through continual learning.


Intuitively, representation forgetting can negatively influence the general knowledge stored in pre-trained LMs, but the relation between them is rarely investigated.  The general knowledge determines how the generality of pre-trained LMs evolves during continual training, i.e. the potential power and capacity of pre-trained LMs for future downstream tasks. 
In this paper, we draw attention to the following fundamental questions of representation forgetting in pre-trained LMs:

\vspace{-0.1mm}

\begin{enumerate}
    \item \textit{Is the generality of the pre-trained LMs  destructed  through representation forgetting?}
     \item \textit{What types of general knowledge of the representation are saliently forgotten?}
     \item \textit{How can we mitigate the general knowledge forgetting of representations?}
\end{enumerate}
\vspace{-0.1mm}

To answer these questions, we carry out empirical studies, with a proposed metric -- overall generality destruction (GD) -- to evaluate the generality evolution, which compares the model performance in continual training with those solely fine-tuned on each task. Given that GD relies on specific tasks, we propose to use probing \cite{tenney2019you,merchant2020happens,zhu2022} to evaluate the general knowledge stored in the pre-trained LMs.
In particular, we propose two more metrics namely syntactic knowledge forgetting (SynF) and semantic knowledge forgetting (SemF) based on probing, to measure the forgetting of syntactic and semantic knowledge, respectively.

We use GD, SynF, and SemF to evaluate the general knowledge forgetting of different pre-trained LMs and different task sequences. Results from extensive experiments show that general knowledge forgetting exists when continually training pre-trained LMs. We further find that the task sequences affect general knowledge forgetting. If general tasks are first learned and application tasks are learned afterward, general knowledge forgetting can be mitigated. We also revisit several continual learning methods and find that generality destruction and general knowledge forgetting vary significantly in different types of learning strategies: (1) although regularization-based methods can maintain the general knowledge injected into pre-trained language models, it restricts the fine-tuning performance of the downstream tasks; (2) rehearsal-based methods suffer from stronger generality destruction and more significant general knowledge forgetting compared with that without strategies; (3) hybrid methods combing rehearsal and regularization achieve more balanced results, achieving low generality destruction while maintaining the general knowledge well.
To our knowledge, we are the first to evaluate general knowledge forgetting 
 in contextualized representations  during continual learning, which can provide a research basis for incremental knowledge injection to pre-trained LMs in a continual learning paradigm.
 

\section{Related Work}
\textbf{Fine-tuning Pre-trained LMs.} It has been proved that pre-training on large-scale unlabelled corpus can inject rich information into LMs and fine-tuning the pre-trained LMs for downstream tasks has become a new learning schema in NLP \cite{devlin2019bert,liu2019roberta,radford2019language}.  Due to the universal language representations in pre-trained LMs, it becomes promising to enable a model to solve various tasks in continual learning scenarios \cite{ke2022continual}. Some methods without fine-tuning are introduced such as in-context learning,  which  induces pre-trained LMs to perform a task by feeding in concatenated and prompted input-target examples along with an unlabeled query example \cite{radford2019language,brown2020language}. But compared with fine-tuning, in-context learning suffers from some practical problems such as  large computation costs  for inference \cite{liu2022few}, uninterpretable behaviors \cite{zhao2021calibrate,min2022rethinking}, and oversensitive to example choices and instruction wording \cite{gao2020making,schick-schutze-2021-just}.


\textbf{Continual learning}. Various continual learning methods are designed to overcome cartographic forgetting \cite{lopez2017gradient,WUTONG,ke2022continual}.   Under fixed model scenarios, regularization-based methods design regularization terms to control the representation drift from previous models \cite{li2017learning,kirkpatrick2017overcoming,aljundi2018memory}. Rehearsal-based methods rely on storing old samples and re-using the samples to recover previous task-related knowledge such as Experiences Replay \cite{de2019episodic,riemer2019learning}.  Some studies also propose hybrid strategies combining regularization and rehearsal such as 
 DERPP \cite{buzzega2020dark}.


From the view of representations, \citet{yu2020semantic} discuss the phenomenon of representation drift in continual learning and propose a method to estimate such drift to mitigate CF. \citet{davari2022probing} further find that even though the model performance on previous tasks is well preserved, the representations still drift due to the parameter update. They measure such representation forgetting by the difference in the performance of optimal linear classifier before and after a new task is introduced.  However, the analysis of representation forgetting is still limited to previously learned tasks. 
In this paper, we step forward to analyze the generality destruction and general knowledge forgetting, which can provide a research basis for injecting general knowledge into pre-trained LMs through continual learning.

\section{Method}
Formally, in the task-incremental continual training setup, a model learns several classification tasks denoted as $\mathcal{T} = \{T^m\}, m = 1,2,...,N$ ($N$ is the length of the task sequence), and each task $T^m \in \mathcal{T}$  contains a limited set of labels $C^m$. For  solving different  tasks, we assume that the label set $C^j\cap C^k = \emptyset, $ if $j \neq k$.  During the training of each task $T^m \in \mathcal{T}$, only the corresponding data $D^m = \{(x^m_i,y^m_i)\}$ are available, where $x^m_i$ is the input text and $y^m_i \in C^m$ is the corresponding label. In the scenario of task-incremental continual training, the task id can be observed when carrying out inference.

\subsection{Probing of General Knowledge}
We define a probing task set $\mathcal{P} = \{P^s\}$.  After  continual training on the specific task $T^m$, we can obtain a trained model $\mathcal{M}^m$. 
Then we freeze the parameters of the trained language encoder and follow \citet{zhu2022} to use a 
classifier (parameterized by a multi-layer perceptron) for probing.
After tuning it on the task $P^s$, we calculate the probing accuracy$S^{m,s}_{l}$ as:
\begin{equation}
    S^{m,s}_{l} = \mathit{Classifer}(\mathcal{M}^m_{l}(P^s)),
\end{equation}
where $l$ indicates the layer index of the pre-trained language encoder in $\mathcal{M}^m$. In pre-trained language models, layers close to the inputs, i.e. lower layers, contain more surface knowledge, and the upper layers contain more syntactic and semantic knowledge \cite{jawahar-etal-2019-bert,tenney-etal-2019-bert}. Therefore, we mainly leverage the probing results of the last layer, denoted as $S^{m,s}_{-1}$

For the probing task set $\mathcal{P}$, we first consider 6 tasks from SentEval \cite{conneau2018senteval}, which has been shown effective to predict the model performance on the downstream task \cite{zhu2022}. In addition, for more salient  implications for semantic knowledge detection, we add the task of MRPC \footnote{\href{https://www.microsoft.com/en-us/download/details.aspx?id=52398}{https://www.microsoft.com/en-us/download/details.aspx ?id=52398.}} from GLUE to further characterize the semantic knowledge of pre-trained language models. MRPC aims to identify whether each sequence pair is paraphrased.  Overall, the above tasks can be approximately divided into two categories -- syntactic and semantic:

\textbf{Syntactic}: bigram shift (BShift), tree depth (TreeDepth), subject number (SubjNum), and object number (ObjNum). We denote the syntactic probing tasks as $\mathcal{P}^{Syn} \subset \mathcal{P}$. 

\textbf{Semantic}: past tense (Tense), MRPC, and coordination inversion (CoordInv). We denote the syntactic probing tasks as $\mathcal{P}^{Sem} \subset \mathcal{P}$. 

Following \citet{zhu2022}, we randomly select 1,200 samples per class in the probing tasks, around 1\% to the original SentEval data. The details of the probing task definitions are shown in Appendix A.

\subsection{Evaluation}
After the model finishes learning on the task $T^m$, we evaluate it on the test set of previous tasks $T^j$ ($j\leq m$), and denote the test accuracy as $R^{m,j}$. We first calculate the average accuracy of each task after training on it  as a surrogate of the overall generality of pre-trained language models:
\begin{equation}
    G = \frac{1}{N} \sum_{m=1}^{N} R^{m,m}
\end{equation}
A stronger value of G means that the model has more power to solve the downstream tasks.
In order to evaluate the generality destruction, we first propose an intuitive metric between the overall generality of a model trained during continual training and that without continual learning:
\begin{equation}
    GD = \frac{1}{N-1} \sum_{m=2}^{N}R^m_* - R^{m,m}
\end{equation} 
where $R^m_*$ is the accuracy of the pre-trained model solely trained on the task $T^m$ \footnote{We adopt a broad classification layer for all the labels in the task for continual learning. To mitigate the effect of model initialization, the results of single training are also carried out with the same layer and sequence of continual tasks. }. 

The metric GD mainly evaluates the surface effect of representation forgetting on previous tasks. To further reveal what general knowledge is forgotten. Thus, we propose the other two metrics to interpret the effect of representation forgetting:
\begin{equation}
    SynF = \frac{1}{|\mathcal{P}^{Syn}|}\sum_{P^s\in \mathcal{P}^{Syn}} \frac{1}{N} \sum_{m=1}^{N} \frac{ S^{*,s}_{-1}-S_{-1}^{m,s}}{S^{*,s}_{-1}} ,
\end{equation}
and 
\begin{equation}
    SemF =\frac{1}{|\mathcal{P}^{Sem}|}\sum_{P^s\in \mathcal{P}^{Sem}} \frac{1}{N} \sum_{m=1}^{N} \frac{S^{*,s}_{-1}-S_{-1}^{m,s}}{S^{*,s}_{-1}},
\end{equation}
where $S^{*,s}_{-1}$ is the result on the probing task $P^s$ using initial pre-trained language model (such as \textit{bert-base-uncased}) as encoder. These two metrics evaluate the evolution of the model's linguistic capacities in semantics and syntactic  after continual learning on downstream tasks.

\section{Tasks and Models}
\subsection{Continual Tasks}
We adopt classification tasks from the benchmark GLUE \cite{wang2018glue} and tasks from MBPA++ \cite{huang2021continual,de2019episodic}. 
Dissimilar tasks are selected to form the continual task sequences, 1) CoLA \cite{warstadt2019neural} requires the model to  determine whether a sentence is linguistically acceptable; 2) MNLI \cite{williams2017broad} is a crowd-sourced collection of 433k sentence pairs annotated with textual entailment information; 3) QNLI\footnote{\href{https://quoradata.quora.com/First-Quora-Dataset-Release-Question-Pairs}{https://quoradata.quora.com/First-Quora-Dataset-Release-Question-Pairs}} requires the model to decide whether the \textit{answer} answers the \textit{question}; 4) QQP (parsed from SQuAD \cite{rajpurkarsquad}) tests the model to correctly output whether a pair of Quora questions are synonymous; 5) Yelp \cite{zhang2015character} requires detecting the sentiment of a sentence; 6)Yahoo \cite{zhang2015character} requires identifying the categorization of questions.

Following \citet{huang2021continual}, we design two types of task sequences: 1) a sequence of 4 classification tasks containing Yahoo, Yelp, MNLI, and CoLA; 2) a sequence of 6 classification tasks containing QQP, Yahoo, MNLI, CoLA, Yelp, and QNLI. The task orders for experiments are shown in Table \ref{sequence} \footnote{According to preliminary experiments, Orders 1 and 2 are the most representative to show the influence of task orders. And without losing generality, we randomly select task sequences for Orders 3-6.}.

\begin{table}[] \small
\centering
\begin{tabular}{cl}
\hline
\multicolumn{2}{c}{Task Orders}          \\ \hline
1      & Yahoo  $\rightarrow$ Yelp $\rightarrow$ MNLI$\rightarrow$ CoLA \\
2      & CoLA $\rightarrow$ MNLI $\rightarrow$ Yelp $\rightarrow$Yahoo  \\
3     & Yelp $\rightarrow$Yahoo $\rightarrow$CoLA$\rightarrow$MNLI \\ 

4 & QQP $\rightarrow$Yahoo $\rightarrow$CoLA$\rightarrow$MNLI $\rightarrow$Yelp$\rightarrow$ QNLI \\
5 & QNLI $\rightarrow$Yelp $\rightarrow$MNLI$\rightarrow$CoLA $\rightarrow$Yahoo$\rightarrow$ QQP \\
6& MNLI $\rightarrow$ Yahoo $\rightarrow$QNLI$\rightarrow$QQP $\rightarrow$Yelp$\rightarrow$ CoLA \\

\hline
\end{tabular}
\caption{Different orders of task sequences for continual learning in experiments.}
\vspace{-5mm}
\label{sequence}
\end{table}


\subsection{Continual Learning Methods}
We mainly revisit        the rehearsal method, Experience Replay (ER) \cite{de2019episodic,robins1995catastrophic}; the regularized-based methods, Learning without Forgetting (LwF) \cite{li2017learning}; the hybrid method, Dark Experience Replay (DERPP) \cite{buzzega2020dark}. The model details are shown as follows:

 \noindent\textbf{ER} \cite{robins1995catastrophic,de2019episodic}. After training the model on the task $T^{m}$, several samples are saved in the memory buffer, and in the following training, the data in the memory buffer are  replayed to recover the knowledge of previous tasks. 

 \noindent\textbf{LwF} \cite{li2017learning}. The method aims to keep the output of new data on the previous label space  to be close to the previously trained model $\mathcal{M}^{m-1}$. 
 For simplicity, we adopt the modified version \cite{yu2020semantic} to keep  the representations:
 \begin{equation}
     \mathcal{L}_{LwF} = ||z_i^{m} - z_i^{m-1}||, 
 \end{equation}
 where $z_i^{m}$ is the representation of $x^m_i$ from the encoder of  $\mathcal{M}^{m}$,  $z_i^{m-1}$ is that from the encoder of  $\mathcal{M}^{m-1}$, and $||\cdot||$ refers to the  Frobenius norm. The training objective is a joint loss of cross-entropy for prediction and the regularization term, i.e. $\mathcal{L} = \mathcal{L}_{ce}+\lambda \mathcal{L}_{LwF}$.

 \noindent\textbf{DERPP} \cite{buzzega2020dark}. The method is a hybrid one combining the regularization strategy and the rehearsal strategy. After training on  the task $T^m$, a limited number of data samples $X^m$ are saved in a memory buffer together with its labels $Y^m$ 
 and output logits  $L^m$. Then during the training of task $T^{m+1}$, a regularization term is used to control the outputs of the current model to be close to the memory buffer.
 It can be expressed as:
 \begin{equation}
 \begin{split}
          L_{DERPP} = \alpha * ||\mathcal{M}^{m+1}(X^m) - L^m||_2 + \\ \beta  * 
          CE (\mathcal{M}^{m+1}(X^m),Y^m) 
 \end{split}
 \end{equation}
 where $\alpha$ and $\beta$ are hyper-parameters, and $CE$ is the cross-entropy loss. The training objective of DERPP is $\mathcal{L} = \mathcal{L}_{ce}+\mathcal{L}_{DERPP}$.

\begin{table}[] \small
\begin{tabular}{lccccc}
\hline
           & Order  & Yahoo & Yelp  & MNLI  & CoLA  \\ \hline
BERT       & Single & 74.75 & 62.33 & 74.14 & 82.34 \\
           & 1 $\rightarrow$     & 0.00  & +0.04 & -1.50 & -0.55 \\
           & 2 $\leftarrow$     & -0.18 & +0.03 & -0.97 & 0.00  \\
\hline
DistilBERT & Single & 74.38 & 61.54 & 70.44 & 79.17 \\
           & 1 $\rightarrow$      & 0.00  & -0.73 & -2.77 & -1.34 \\
           & 2 $\leftarrow$     & -0.62 & -0.81 & -1.12 & 0.00  \\
 \hline
ALBERT & Single & 73.17	&61.62&	72.41&	81.16\\
           & 1 $\rightarrow$ &0.00& -0.69 & -2.15&-6.53  \\
           & 2 $\leftarrow$& +0.10&-1.16&-1.23&0.00      \\
 \hline
RoBERTa    & Single & 74.86   & 65.51    & 82.11    & 84.77    \\
           & 1  $\rightarrow$    & 0.00 & -1.12 & -2.18 & -2.08 \\
           & 2  $\leftarrow$    & -0.18  & -0.60 & -1.29 & 0.00   \\
\hline 
\end{tabular}   
\caption{Accuracies of different pre-trained models on single tasks and continual tasks of Order 1 and 2.  Each line of these two Orders is the relative change compared with fine-tuning on the \textit{single} task.} 
    \vspace{-4.5mm}
\label{exp1}
\end{table}

\begin{table*}[] \small
\centering
\begin{tabular}{c|p{0.03\textwidth}p{0.03\textwidth}p{0.03\textwidth}p{0.04\textwidth}|p{0.03\textwidth}p{0.03\textwidth}p{0.03\textwidth}p{0.04\textwidth}|p{0.03\textwidth}p{0.03\textwidth}p{0.03\textwidth}p{0.04\textwidth}|p{0.03\textwidth}p{0.03\textwidth}p{0.03\textwidth}p{0.04\textwidth}}
\hline
   & \multicolumn{4}{c|}{BERT}                                                                     & \multicolumn{4}{c|}{DistilBERT}                                                              & \multicolumn{4}{c|}{ALBERT}                                                                   & \multicolumn{4}{c}{RoBERTa}        \\
              & G & GD                   & SynF                 & SemF                 & G                  & GD & SynF                 & SemF                 & G & GD                    & SynF                 & SemF                 & G & GD                   & SynF                 & SemF                 \\ 
              \hline
1              &        73.39                &      0.67                &       6.50               &  0.72                  &     71.38                 &   1.61                     &   18.83                   &  9.95                    &           68.25              &  3.12                    &   19.67                   & 10.50                     &            75.47             &   1.79                   &     19.98                 &        11.32              \\
2              &         73.41                &    0.37                  &    2.96                  &     -1.00                 &    71.21                  &    0.85                    &     15.67                 &   9.32                   &        72.13                 &   0.76                   &    11.95                  &    7.37                  &                76.30         &   0.69                   &     17.62                 &     8.42                 \\
3              &        72.69                 &     0.92                 &      5.26                &    1.53                  &  70.32                    &     1.41                   &      18.32                &     11.12                 &        69.88                 &  2.95                    &      20.65                &     10.35                 &          76.17               &  0.85                    &      18.00                &      9.67                \\
4             &      73.54                   &   0.73                   &    5.62                  &    1.75                  &         71.50            &  0.96                      &   16.33                   &    8.89                  &      69.89                   &   3.29                   &   21.78                &    12.95                 &           76.50               &    1.00                  &    17.77                  &   10.45                   \\
5                    &   73.64                       &   0.61                   &     4.72     &   0.65                 &       70.84               &     1.66                   &      19.34                 &   10.03                   &     72.51                   &  1.65                    &    17.06                  &      11.84                &       76.78                  &     0.88                 &     17.17                &   8.68                   \\
6                    &   73.57                      &  0.66                    &  5.73                   &   0.94                 &     71.60                 &    0.80                    &    13.21                 &   9.04                   &     71.32                    &   2.84                   &   19.30                   &  10.84                    &          76.01               &     1.43                 &    19.88                  &   9.94                   \\

\hline
\end{tabular}
\caption{Results of continual learning on different language models and different task sequences. The left column is the order of the task orders corresponding to Table \ref{sequence}. All the models are trained with three different and fixed random seeds, and average results are reported. For probing tasks, results are averaged with five different runs.}
\label{main}
\vspace{-5mm}
\end{table*}

\section{Experiments and Results}
 We first show the generality of pre-trained language models is destructed for representation forgetting through an empirical study in Section 5.1. Then we evaluate generality destruction and  general knowledge forgetting for different pre-trained LMs and different task orders in Section 5.2. And we further analyze the probing results at two representative task orders to show the effect of task sequences in Section 5.3. Finally, we test the model with different continual learning strategies to show the generality evolution in explicit controls of the representations in Section 5.4.

\subsection{Implementation Details}
Following \citet{zhu2022}, we use several wide-use pre-trained language modeling for evaluation such as BERT \cite{devlin2019bert}, DistilBERT \cite{sanh2019distilbert}, ALBERT \cite{lan2019albert}, and RoBERTa \cite{liu2019roberta}. The details of the pre-trained language models are introduced in Appendix B. We all use the base version of pre-trained language models provided by HuggingFace \footnote{\href{https://huggingface.co/}{https://huggingface.co/}}. 
 We train our model on 1 GPU (Tesla V100 32G) using the Adam optimizer \cite{kingma2014adam}. For all the models, the batch size is 32,  the learning rate is 3e-5, and the scheduler is set linear.  The max sequence length of the inputs is 512. We train our model 10 epochs with the early stopping of patience 20. The results are all averaged in different random seeds.

\subsection{Generality Destruction}
 We first show an empirical study of continual training to demonstrate the destruction of generality caused by representation forgetting. The  continual learning example contains Orders 1 and 2 on different pre-trained language models in Table \ref{exp1}. We compare the model performance $R^{m,m}$ ($T^m \in \mathcal{T}$) with the model performance solely trained on each task $R_{*}^m$ (for simplicity, we call the difference $R^{m,m}-R_{*}^m$ as $R^{-}$ in the task $T^m$).

As can be observed in Table 3, the results of $R^{-}$ in mostly all tasks are negative, showing that all continually trained models  suffer from decreases in performance compared with those trained on a single task. This phenomenon demonstrates that the generality of the pre-trained language models is destructed, and the degree varies in different models.
In Order 2 the results of $R^-$ are mostly smaller than those in Order 1.   For example, $R^-$ are -1.50\%, -2.77\%, -2.15\%, and -2.18\% in the task MNLI in Order 4 for BERT, DistilBERT, ALBERT, and RoBERTa, respectively, but are only -0.97\%, -1.12\%, -1.23\%, and -1.29\% in Order 2. It implies that the order of tasks can affect the degree of generality destruction, which we further analyze in Section~\ref{sec:detailedprobing}.

\subsection{Evaluation of Representation Forgetting}
We show the continual learning results of four pre-trained language models and different task orders in Table \ref{main}. For the overall generality G,  RoBERTa achieves the strongest performance, 75.47\%, 76.30\%, 76.17\%, 76.50\%, 76.78\%, and 76.01\%  for Order 1-6, respectively. And the worst is ALBERT, only 68.25\%, 72.13\%, 69.88\%, 69.89\%, 72.51\%, and 71.32\%, respectively. Such results indicate that the stored general knowledge is the richest in RoBERTa, but it is the most scarce in ALBERT for the smaller size of parameters. 

The results of GD are all positive in the  experiments, indicating that all models suffer from general knowledge forgetting in representations.
BERT has the strongest ability to maintain general knowledge and results in the lowest GD.
Although RoBERTa has a strong generality, the model suffers from more forgetting of general knowledge compared with BERT. The most fragile model is ALBERT, achieving  3.12\%, 0.76\%, 2.95\%, 3.29\%, 1.65\%, and 2.84\% GD values. Interestingly, the GD values of Order 2 are mostly the lowest  for all pre-trained language models, 0.37\%, 0.85\%, 0.76\%, 0.69\% for BERT, DistilBERT, ALBERT, and RoBERTa, respectively. The phenomenon indicates that the sequence of tasks has a strong effect on general knowledge forgetting.

For SynF and SemF, the results can also reflect the general knowledge forgetting from the view of syntactic and semantics. We calculate the correlation between GD, SynF, and SemF. The Pearson correlation between GD and SynF is 0.63, that between GD and SemF is 0.67, and that between SynF and SemF is 0.97, where the significance level $ p<0.01 $. It shows that the results of SynF and SemF are positively correlated to GD, indicating the forgetting of syntactic and semantic knowledge can reflect generality forgetting. SynF and SemF show a strong positive correlation with each other, indicating that syntactic and semantic knowledge are generally forgotten together.  Similar to GD, BERT also has the strongest ability to overcome general knowledge forgetting compared with other pre-trained models (the SynF and SemF are all below 10.0\% in the tasks), and ALBERT is the most fragile. 

\begin{figure*}
    \centering
    \includegraphics[width=0.95\hsize]{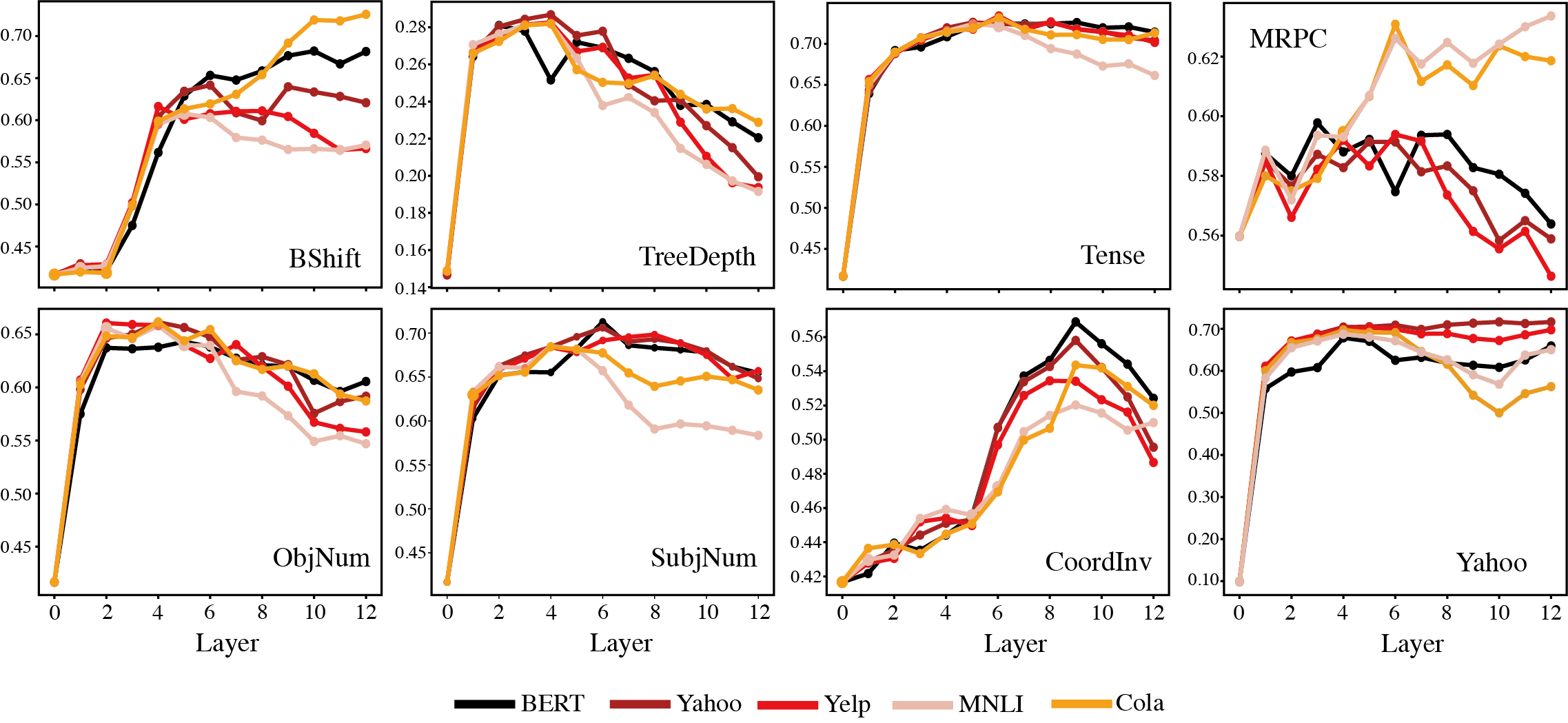}
    \caption{Probing results on BERT through continual training on Order 1. BERT refers to the initialized \textit{bert-base-uncased}, and others are the probing performance after training on the corresponding task.}
    \label{probing1}
    \vspace{-5mm}
\end{figure*}

\subsection{Detailed Probing Results}
To further analyze the influence of each task during continual training, we  empirically illustrate the detailed probing results of the BERT encoder in Order 1 and 2.  Overall, we find that general knowledge and injected world knowledge are forgotten step by step for application tasks, but the general knowledge can be maintained through learning on general linguistic tasks.

\label{sec:detailedprobing}
\subsubsection{Continual Learning on Order 1}
The detailed probing results of the BERT encoder in Order 1 in shown in Figure \ref{probing1}. To further analyze the influence of each task during continual training, we  empirically illustrate the detailed probing results of the BERT encoder in Order 1 (shown in Figure \ref{probing1}).  The results include probing accuracy on all 7 probing tasks from all 12 layers of the BERT encoder (among which Bshift, ObjNum, TreeDepth, and SubjNum are syntactic tasks, Tense, CoordInv and MRPC are semantic tasks). We also use probing to test the BERT encoder knowledge of the initial task Yahoo, which can be regarded as injected world knowledge in pre-trained LMs. Overall, we observe that the change of the probing results mainly happens on the upper layers (6-12) of BERT, which demonstrates that fine-tuning has a more significant influence on the upper layers (similar to \citet{WUTONG}).

Most probing results decrease increasingly with the training on Yahoo, Yelp, and MNLI, which means the general knowledge is forgotten step by step.
Then the tasks of Yahoo and Yelp are specific downstream tasks for applications, and they cause a significant decrease in all seven probing tasks. It implies that to fit the representations for specific semantic-related downstream tasks, general  knowledge is regarded as useless and discarded. The probing results of Tense, MRPC, and CoordInv decrease less significantly, indicating that semantic information is less sensitive than syntactic information and is mostly maintained. The performance of the probing task MRPC is boosted and outperforms that of the initial BERT encoder because MNLI requires the model to learn the entailment relation between sentences, similar to MRPC.  The task  CoLA, which is a general linguistic task to learn whether a sentence is in acceptable grammaticality, helps recover the syntactic knowledge and semantic knowledge of the BERT encoder. In some probing tasks such as BShift and TreeDepth, the probing results even surpass the initial BERT encoder, which means richer knowledge can be injected into the representations.

Finally, for the probing results of Yahoo, we observe that the knowledge of document classification is incorporated into the representations after training on the Yahoo task, and such knowledge is maintained after training on Yelp (higher probing performance than that of initial BERT). The results demonstrate that although the performance of a task decreases after training on the newly arriving task, it is possible to maintain the incorporated knowledge in representations. But the probing results are lower than those of initial BERT after training on CoLA,  which implies different learning tasks have different effects on the incorporated knowledge.

\subsubsection{Continual Learning Order 2}
One interesting observation is that in Table 2 and Figure 2, GD of Order 1 is significantly lower than those of others.  Thus, we also illustrate the probing results of Order 2, where CoLA is first trained (in Figure \ref{probing2}).
After training on the task of CoLA, the results of all seven probing tasks increase significantly on the upper layers, and all outperform those of initial BERT representations. The results are also higher than  those of continual learning after CoLA in Order 1. It indicates that CoLA, which is a general linguistic task, can inject general knowledge into pre-trained language models.  But with the continual learning of MNLI, Yelp, and Yahoo, the probing results decrease step by step, and they are primarily below initial BERT (the red, pink, and orange curves in Figure 3 are all below the black curve). 


 \begin{figure*}
    \centering
    \includegraphics[width=0.95\hsize]{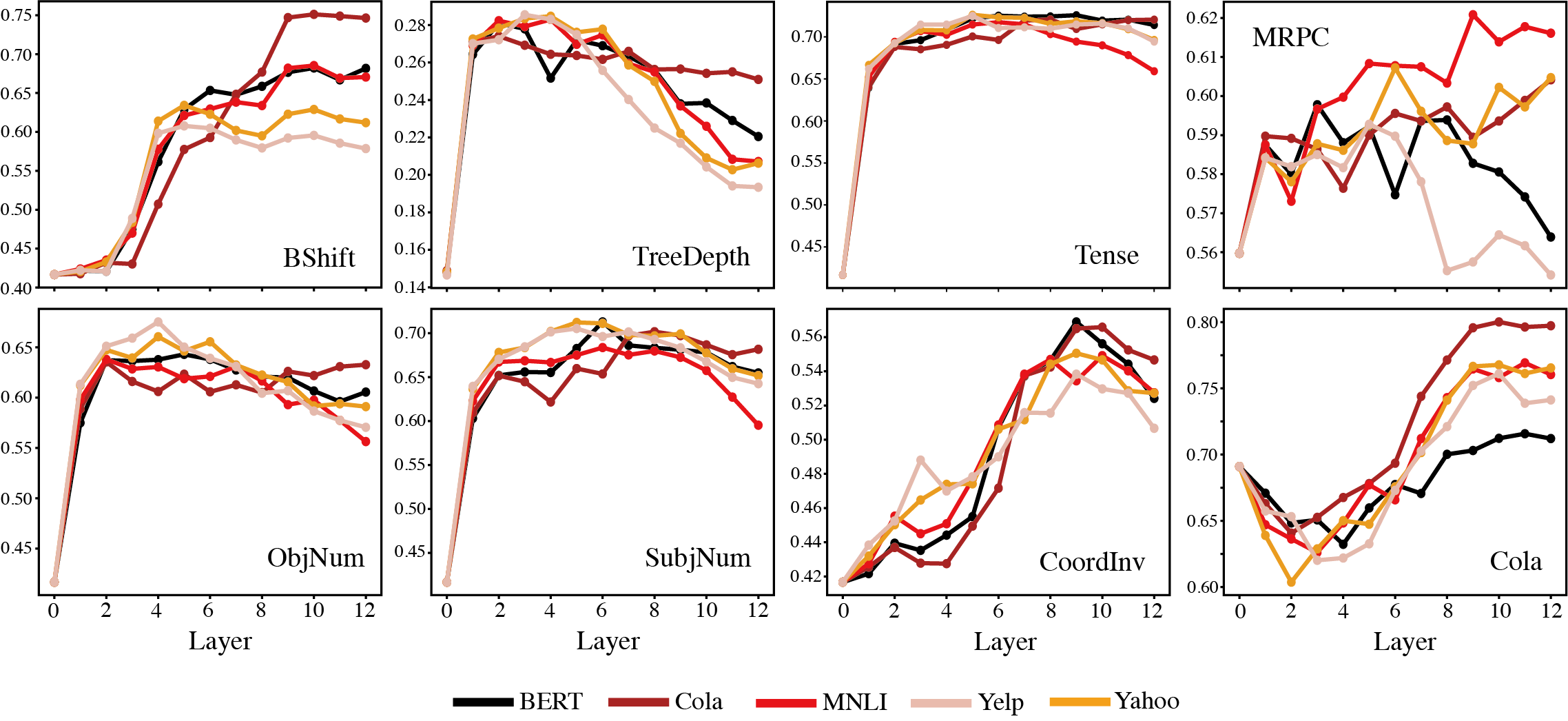}
    \caption{Probing results on BERT through continual training on  Order 2, which is a reverse sequence of Order 1.}
    \label{probing2}
    \vspace{-2mm}
\end{figure*}

 \begin{figure*}
    \centering
    \includegraphics[width=\hsize]{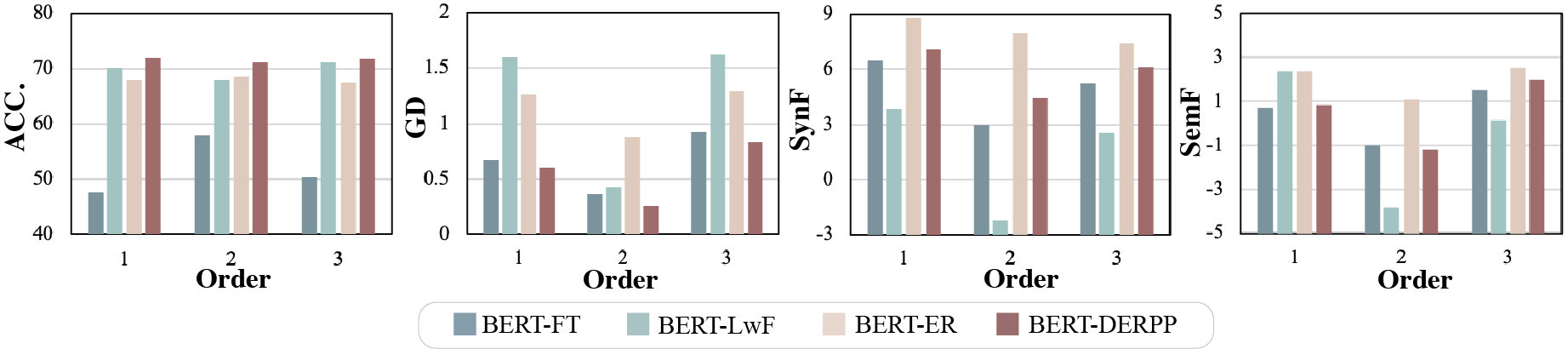}
    \caption{Evaluations of different continual learning strategies (FT refers to continual learning without any strategy).}
    \label{probing2}
    \vspace{-5mm}
\end{figure*}

For the world knowledge of CoLA, the probing results decrease after training on the task of MNLI, Yelp, and Yahoo, showing that the knowledge of CoLA is forgotten to some extent due to representation forgetting.
But the result curves of trained models are all above the curve of initial BERT, which implies the knowledge of grammar is injected into BERT and well-maintained after continual training on the other tasks. 

\subsection{Different Continual Learning Methods}
We show the results of different continual learning strategies on BERT in Figure 4. Besides the results of GD, SynF, and SemF, we also show the average performance on all tasks for each order after finishing continual learning. For example, the model BERT-FT only achieves  47.75\%, 57.9\%, 50.25\%, 49.42\%, 43.33\%, and 50.91\% for  Order 1-6, respectively. But BERT-ER achieves 67.62\%, 68.53\%, 68.75\%,  69.01\%, 69.18\%, and 64.71\%, correspondingly. The best average performance is achieved by the method BERT-DERPP, and the worst is that of BERT-FT. 
The average performance indicates the effectiveness of continual learning strategies to maintain performance on previously trained tasks. 

We observe that GD of BERT-LwF is all higher than those of BERT-FT, implying regularization-based methods suffer from stronger generality destruction. It 
can be explained that by maintaining the representations close to the previously learned ones, the model is limited to fit the downstream tasks. The results show that it is a trade-off problem for regularization methods to balance the performance of fine-tuning on downstream tasks and representation maintenance. It also implies that in such a representation-controlled situation, GD can not effectively reflect the change of generality.  But the SynF and SemF of BERT-LwF are always lower than those of BERT-FT, indicating that the general knowledge injected into the model can be well maintained, (SynF and SemF are even negative (-2.24\% and -3.84\%)).

For the rehearsal-based method -- BERT-ER, we find that the models suffer from more drastic generality destruction compared with BERT-FT. BERT-ER achieves the GD of 1.6\%, 0.42\%, and 1.63\% on Order 1-3, which are 0.93\%, 0.05\%, and 0.71\% higher than those of BERT-FT, respectively. At the same time, SynF and SemF of BERT-ER are also higher than those of BERT-FT. It indicates that the generality of the model is destructed and general knowledge is forgotten more severely compared with BERT-FT.

Compared with the above methods, the model BERT-DERPP achieves the best ACC, and the lowest GD values (even lower than BERT-FT), which implies that DERPP is capable to mitigate the problem of  catastrophic forgetting  and generality destruction.  Meanwhile, general knowledge forgetting is also acceptable, i.e. SynF values are all lower than BERT-ER and SemF values are mostly lower than BERT-FT. The results also show that the hybrid model can also  preserve more  general knowledge compared with those solely considering rehearsal and regularization.

We also find that in the experiments of continual learning strategies, GD, SynF, and SemF are still strongly related to the task order, which is also observed in Section 5.3. They are the lowest in Order 2 and are mostly the highest in Order 3. 
 The results further demonstrate  that training on the general linguistic task first can help preserve more general knowledge when using different continual learning strategies. GD is also positively correlated to the SynF and SemF, which shows a significant relationship between them, which shows that generality destruction can also be reflected by general knowledge forgetting based on probing results.



\section{Conclusion}
In this paper, we investigated the effect of representation forgetting on 
 the generality of pre-trained LMs through continual learning. Experiments showed that the general knowledge in the representations is negatively  forgotten during  continual training, destructing the generality of the pre-trained language models.
 The extensive results provided us with several insights into the generality destruction caused by representation forgetting, which could serve as a research basis for injecting general knowledge into pre-trained language models through continual learning.
 

\section{Limitations}
In this study, we mainly focus on giving a post-hot analysis of the effect of representation forgetting in continual learning, which is a promising learning paradigm for incremental knowledge injection into pre-trained LMs. Our analysis gives a new perspective on model generality for knowledge forgetting.
But the method may be less effective for analyzing the seq2seq models, and the in-context learning paradigm. In future work, we will further investigate the knowledge-forgetting problem in in-context learning, which is an emergent and effective paradigm in the increasingly large seq2seq language models.

\bibliography{ref}
\bibliographystyle{acl_natbib}
\clearpage
\appendix

\section{Probing Task Details}
\label{sec:appendix}

We show the definitions of the probing tasks as allows:

\begin{enumerate}
    \item \noindent\textbf{BShift}: predicting whether two consecutive tokens within the sentence have been inverted).
    \item \noindent\textbf{Tree Depth}:  predicting the maximum depth of the sentence's syntactic tree.
\item \noindent\textbf{Coordination Inversion}:  distinguishing whether the order of two coordinated clausal conjoints has been inverted.

\item \noindent\textbf{Subj Number}: focusing on the number of the subject of the main clause.

\item \noindent\textbf{Obj Number}: focusing on the number of the direct object of the main clause.
\item \noindent\textbf{Past Present}: predicting whether the main verb of the sentence is marked as being in the present (PRES class) or past (PAST class) tense.
\item \noindent\textbf{MRPC}: indicating whether each pair captures a paraphrase/semantic equivalence relationship.
\end{enumerate}

\section{Pre-trained Language Models}
\noindent \textbf{\textrm{BERT}} \cite{devlin2019bert} is the most representative model that leverages the deep bi-directional transformers as the backbone network. It pre-trains the model by carrying out Masked Language Model (MLM) and the Next Sentence Prediction (NSP) as the self-supervision tasks. 

\vspace{2mm}

\noindent \textbf{\textrm{DistilBERT}} \cite{sanh2019distilbert} leverage the method of knowledge distillation during the pre-training phase. DistilBERT introduces a triple loss containing language modeling, distillation, and cosine-distance losses.   It can retain 97\% of language understanding capacities but with only 60\% size of a BERT model.

\vspace{2mm}

\noindent \textbf{\textrm{ALBERT}} \cite{lan2019albert}  is a lite version of BERT, which share the parameters of Transformers blocks across all layers, and the embeddings matrics are factorized into two small size ones. ALBERT can achieve close performance to that of BERT by fine-tuning the downstream  task.

\vspace{2mm}

\noindent \textbf{\textrm{RoBERTa}} \cite{liu2019roberta} pre-trains the model BERT on over 160GB English corpora by adopting improved techniques such as dynamic sampling, large mini-batches, and masked language modeling without the task of next sentence prediction.

\end{document}